\documentclass[sigconf]{acmart}
\setcopyright{none}
\renewcommand\footnotetextcopyrightpermission[1]{}
\usepackage{multirow}
\usepackage{booktabs}
\usepackage{amsmath}
\begin{document}

\title{DynaBridge: Dynamic Summary-Guided Cross-Task Multimodal Fusion for DASS-Structured Mental Health Assessment}

\author{Shiyu Teng}
\orcid{0000-0001-7233-5027}
\affiliation{%
  \department{College of Information Science and Engineering}
  \institution{Ritsumeikan University}
  \city{Ibaraki}
  \state{Osaka}
  \country{Japan}
}
\email{tsy082@fc.ritsumei.ac.jp}

\author{Haichen Yu}
\affiliation{%
  \department{College of Information Science and Engineering}
  \institution{Ritsumeikan University}
  \city{Ibaraki}
  \state{Osaka}
  \country{Japan}
}
\email{is0760hh@ed.ritsumei.ac.jp}

\author{Jiaqing Liu}
\affiliation{%
  \department{College of Information Science and Engineering}
  \institution{Ritsumeikan University}
  \city{Ibaraki}
  \state{Osaka}
  \country{Japan}
}
\email{liu-j@fc.ritsumei.ac.jp}

\author{Hao Sun}
\affiliation{%
  \department{College of Information Science and Engineering}
  \institution{Ritsumeikan University}
  \city{Ibaraki}
  \state{Osaka}
  \country{Japan}
}
\email{sunhaoxx@zju.edu.cn}

\author{Yu Song}
\affiliation{%
  \department{College of Information Science and Engineering}
  \institution{Ritsumeikan University}
  \city{Ibaraki}
  \state{Osaka}
  \country{Japan}
}
\email{yusong@fc.ritsumei.ac.jp}

\author{Shurong Chai}
\affiliation{%
  \department{College of Information Science and Engineering}
  \institution{Ritsumeikan University}
  \city{Ibaraki}
  \state{Osaka}
  \country{Japan}
}
\email{is0538kr@ed.ritsumei.ac.jp}

\author{Ruibo Hou}
\affiliation{%
  \department{College of Information Science and Engineering}
  \institution{Ritsumeikan University}
  \city{Ibaraki}
  \state{Osaka}
  \country{Japan}
}
\email{is0539vv@ed.ritsumei.ac.jp}

\author{Lanfen Lin}
\affiliation{%
  \department{College of Computer Science and Technology}
  \institution{Zhejiang University}
  \city{Hangzhou}
  \state{Zhejiang}
  \country{China}
}
\email{llf@zju.edu.cn}

\author{Yen-Wei Chen}
\authornote{Corresponding author.}
\affiliation{%
  \department{College of Information Science and Engineering}
  \institution{Ritsumeikan University}
  \city{Ibaraki}
  \state{Osaka}
  \country{Japan}
}
\email{chen@is.ritsumei.ac.jp}

\renewcommand{\shortauthors}{Anonymous Authors}
\begin{abstract}
Multimodal behavioral analysis offers a scalable approach to assessing depression, anxiety, and stress, yet generic fusion models often ignore the psychometric structure of questionnaire labels. In DASS-21, risk labels are derived from ordered symptom items through fixed item-to-subscale mappings. We propose \textbf{DynaBridge}, a dynamic summary-guided cross-task multimodal framework for DASS-structured mental health assessment. DynaBridge encodes acoustic, visual, and textual cues across multiple sessions and augments them with frozen-LLM-generated DASS-aware summaries as participant-level semantic evidence. It predicts ordinal item distributions, reconstructs depression, anxiety, and stress risk evidence from item-level soft scores, and fuses this evidence with direct multimodal risk predictions. A confidence-aware refinement strategy further incorporates high-confidence semantic cues conservatively. On the official AdoDAS validation split, DynaBridge outperforms the official baseline and representative multimodal methods, achieving 0.5012 mean F1 for D/A/S risk prediction and 0.3216 mean QWK for DASS-21 item prediction. These results show the value of bridging multimodal cues, semantic summaries, and DASS-21 psychometric structure.

\end{abstract}



\maketitle

\section{Introduction}

Depression, anxiety, and stress affect daily functioning, academic performance, and long-term well-being~\cite{who2022world}. Standardized questionnaires and interviews remain central to psychological assessment, but they are episodic, subjective, and difficult to scale. This has motivated automatic mental health assessment from multimodal behavioral cues such as speech, facial behavior, body movement, and language use~\cite{gratch2014distress,ringeval2019avec}.

Recent multimedia and affective computing methods usually integrate acoustic, visual, and textual representations through feature fusion, decision fusion, attention, or multimodal Transformers~\cite{zadeh2017tensor,tsai2019multimodal,hazarika2020misa}. Although effective, most methods formulate assessment as ordinary classification or regression over final labels. This ignores an important property of questionnaire-derived labels: in the Depression Anxiety Stress Scales (DASS), depression, anxiety, and stress scores are computed from fixed subsets of ordinal symptom items~\cite{lovibond1995manual,antony1998psychometric}. Treating item responses and subscale risks as independent outputs can therefore produce inconsistent predictions.

Large language models (LLMs) offer useful semantic support because they can summarize fragmented free responses and organize symptom-related evidence~\cite{brown2020language,openai2023gpt4}. However, directly using LLMs as mental health label predictors is risky: they may over-infer psychological states from short or ambiguous text and may not align with audio-visual cues or psychometric rules~\cite{ji2023survey}. We therefore use frozen LLMs only as auxiliary semantic evidence generators; their summaries never access ground-truth DASS labels.

We propose \textbf{DynaBridge}, a dynamic summary-guided cross-task multimodal framework for DASS-structured mental health assessment. DynaBridge connects multimodal behavioral cues, LLM-generated psychological summaries, DASS-21 item-level ordinal predictions, and depression/anxiety/stress risk assessment. It generates progressive DASS-aware summaries from free-response sessions, embeds them as a participant-level semantic modality, reconstructs risk evidence from item-level soft probabilities using the predefined DASS mapping, and refines item predictions with confidence-aware semantic evidence. Our main contributions are:

\begin{itemize}
    \item We formulate multimodal DASS assessment as a structured ordinal cross-task problem that explicitly connects DASS-21 item responses with depression, anxiety, and stress risk prediction.
    \item We propose DynaBridge, which integrates acoustic, visual, textual, and LLM-generated semantic evidence across multiple psychological response sessions.
    \item We introduce DASS-aware item-to-risk reconstruction and confidence-aware item refinement to improve both predictive performance and item-risk consistency.
\end{itemize}
\section{Related Works}
\subsection{Multimodal Mental Health Assessment}

Automatic mental health assessment from behavioral signals has been studied in multimedia, affective computing, and clinical interview analysis. Resources such as DAIC/DAIC-WOZ and the AVEC series provide audio-visual recordings, transcripts, and questionnaire annotations for psychological distress and depression severity estimation~\cite{gratch2014distress,ringeval2019avec}. Building on these benchmarks, prior work has used acoustic cues, facial behavior, head pose, gaze, language content, and their combinations for depression recognition and severity estimation~\cite{stepanov2017depression,qureshi2019verbal,dham2017depression}.

A key challenge is multimodal fusion. Classical approaches use feature- or decision-level fusion, while recent methods model cross-modal interaction with attention and Transformer-based architectures. Representative methods include Tensor Fusion Network~\cite{zadeh2017tensor}, Low-rank Multimodal Fusion~\cite{liu2018efficient}, Multimodal Transformer~\cite{tsai2019multimodal}, MISA~\cite{hazarika2020misa}, multimodal adaptation gates~\cite{rahman2020integrating}, CubeMLP~\cite{sun2022cubemlp}, and TensorFormer~\cite{sun2023tensorformer}. Emotion- and sentiment-aware models further show that affective context can improve depression detection~\cite{teng2024intra,teng2026prefix,teng2024multi}. However, most methods predict final scores or categories directly. In contrast, DynaBridge exploits the DASS-21 item-to-subscale structure and reconstructs risk predictions from ordinal item-level evidence.

\subsection{LLM-Assisted Psychological Understanding}

LLMs have recently been used for psychological text understanding, including interview summarization, symptom evidence extraction, and mental health-oriented reasoning. LLM-generated prompts or summaries can complement raw transcript embeddings by turning sparse free responses into higher-level semantic evidence~\cite{teng2025emotionprompts,teng2026depressionllm,teng2025cot,hou2025rag,teng2026dynamic}. Nevertheless, LLMs should not be treated as standalone diagnostic predictors because their judgments may be weakly grounded in observable behavior and validated scales. DynaBridge therefore uses LLMs only as frozen evidence summarizers. Unlike prior LLM-assisted depression detection, it integrates global, domain-level, and item-level DASS summaries with multimodal representations and DASS-structured item-to-risk reconstruction.

\section{Method}
\label{sec:method}

\begin{figure*}[t]
\centering
\includegraphics[width=\textwidth]{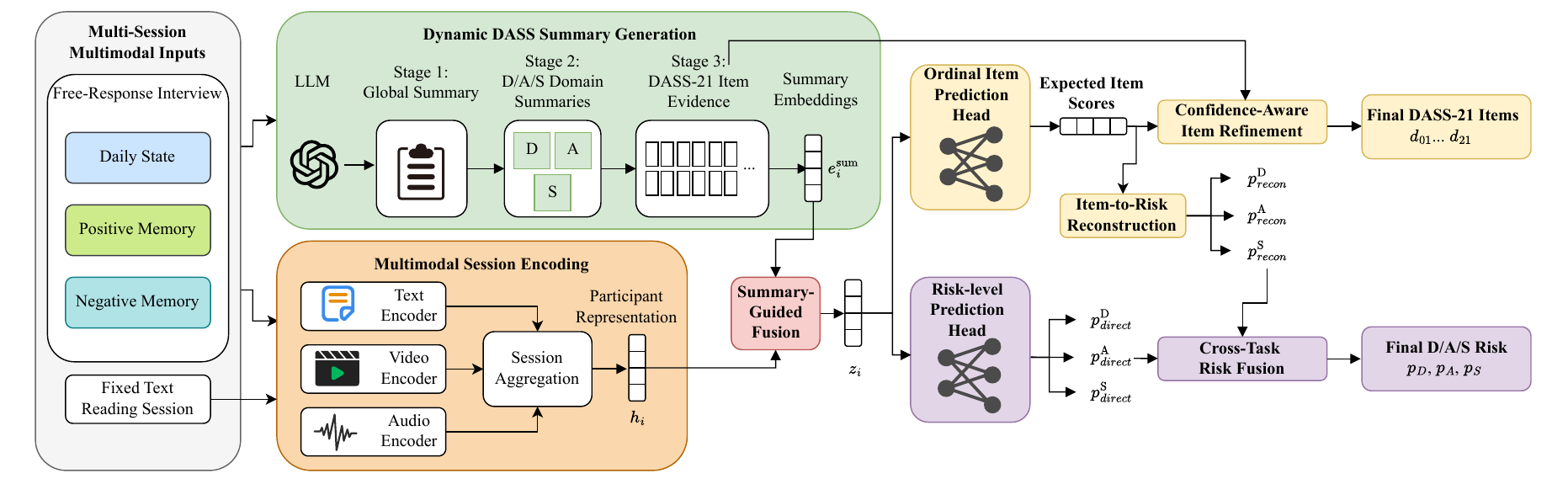}
\caption{Overview of the proposed DynaBridge framework.}
\label{fig:framework}
\end{figure*}

We propose \textbf{DynaBridge}, a dynamic summary-guided cross-task multimodal framework for DASS-structured mental health assessment. DynaBridge bridges multimodal behavioral representations, LLM-generated psychological summaries, DASS-21 item-level ordinal prediction, and depression/anxiety/stress risk assessment. Instead of treating item responses and risk labels as independent targets, it reconstructs risk-level evidence from item-level soft predictions using the predefined DASS-21 structure. Fig.~\ref{fig:framework} gives an overview.

\subsection{Problem Formulation}
\label{subsec:problem}

For participant $i$, the input contains multiple response sessions:
\begin{equation}
X_i = \{X_i^1, X_i^2, \ldots, X_i^T\},
\end{equation}
where each session provides audio and visual observations, and the three free-response sessions additionally provide transcripts. In AdoDAS, $T=4$, with one fixed reading session $t_0$ and three free-response sessions $\mathcal{F}=\{t_1,t_2,t_3\}$. The first objective is risk-level prediction:
\begin{equation}
\mathbf{y}^{\mathrm{risk}}_i = (y_i^D, y_i^A, y_i^S),
\end{equation}
where $y_i^r \in \{0,1\}$ denotes elevated risk in dimension $r \in \{D,A,S\}$. The second objective is DASS item prediction:
\begin{equation}
\mathbf{y}^{\mathrm{item}}_i = (d_{i1}, d_{i2}, \ldots, d_{i21}),
\end{equation}
where each item response $d_{ij}\in\{0,1,2,3\}$ is ordinal. DASS-21 defines fixed item groups:
\begin{equation}
\begin{aligned}
\mathcal{D} &= \{3,5,10,13,16,17,21\},\\
\mathcal{A} &= \{2,4,7,9,15,19,20\},\\
\mathcal{S} &= \{1,6,8,11,12,14,18\}.
\end{aligned}
\end{equation}
This structure allows item-level predictions to reconstruct risk-level scores, making DASS assessment a structured ordinal cross-task problem.

\subsection{Multimodal and Summary-Guided Representation}
\label{subsec:representation}

For each participant, DynaBridge encodes multimodal information at the session level. The AdoDAS protocol contains one fixed reading session and three free-response sessions. The fixed reading session is used only through acoustic and visual cues, since its text content is identical across participants and does not provide participant-specific semantic information. In contrast, the three free-response sessions provide acoustic, visual, and textual information.

Let $\mathbf{a}_{i,t}$ and $\mathbf{v}_{i,t}$ denote the acoustic and visual representations of participant $i$ in session $t$, respectively. For a free-response session $t$, we additionally denote its textual representation by $\mathbf{x}_{i,t}$. The session representation for each free-response session is computed as
\begin{equation}
\mathbf{s}_{i,t}
=
F_{\mathrm{avx}}(\mathbf{a}_{i,t}, \mathbf{v}_{i,t}, \mathbf{x}_{i,t}),
\quad t \in \mathcal{F},
\end{equation}
where $\mathcal{F}=\{t_1,t_2,t_3\}$ denotes the set of free-response sessions. For the fixed reading session $t_0$, only acoustic and visual features are used:
\begin{equation}
\mathbf{s}_{i,t_0}
=
F_{\mathrm{av}}(\mathbf{a}_{i,t_0}, \mathbf{v}_{i,t_0}).
\end{equation}
Both $F_{\mathrm{avx}}(\cdot)$ and $F_{\mathrm{av}}(\cdot)$ project their inputs into the same hidden space. The fixed-reading and free-response session representations are then aggregated into a participant-level representation:
\begin{equation}
\mathbf{h}_{i}
=
F_s(\mathbf{s}_{i,t_0}, \mathbf{s}_{i,t_1}, \mathbf{s}_{i,t_2}, \mathbf{s}_{i,t_3}),
\end{equation}
where $F_s(\cdot)$ denotes the session aggregation function. When school/class identifiers are available at test time, they are encoded as lightweight context priors and concatenated with $\mathbf{h}_{i}$.

DynaBridge further generates DASS-aware semantic summaries only from the three free-response transcripts. Let $C_i$ denote the concatenated transcripts of the daily-state, positive-memory, and negative-memory responses. The frozen LLM never accesses ground-truth DASS-21 item scores, D/A/S risk labels, model predictions, or validation feedback. Summary generation follows a fixed evidence-oriented three-stage protocol:
\begin{equation}
S_i^{(1)} = G_1(C_i),
\end{equation}
\begin{equation}
S_i^{(2)} = G_2(C_i, S_i^{(1)}),
\end{equation}
\begin{equation}
S_i^{(3)} = G_3(C_i, S_i^{(1)}, S_i^{(2)}).
\end{equation}
The first-stage summary captures global participant-level affective and behavioral cues across the free-response sessions. The second-stage summary organizes transcript-supported evidence into depression-, anxiety-, and stress-related categories. The third-stage summary extracts item-level evidence for the 21 DASS items. For each item, the summary records whether the evidence is explicit, weak, or not observed according to the available transcript evidence. When explicit evidence is available, the item-level summary also provides a confidence score and an ordinal cue for subsequent conservative refinement.

To reduce label leakage and hallucinated evidence, the same evidence-oriented prompt templates are used for all participants. The LLM is instructed to describe only transcript-supported cues, mark unsupported symptoms as not observed rather than infer them, and avoid assigning final DASS scores, D/A/S risk labels, or diagnostic conclusions. All summaries are generated before neural model training and cached for both training and inference. In this way, the summary branch serves as a structured semantic evidence extractor rather than a pseudo-label generator.

The three summaries are encoded into a semantic representation:
\begin{equation}
\mathbf{e}_{i}^{\mathrm{sum}}
=
E_{\mathrm{sum}}(S_i^{(1)}, S_i^{(2)}, S_i^{(3)}),
\end{equation}
and fused with the participant-level multimodal representation:
\begin{equation}
\mathbf{z}_{i}
=
F_{\mathrm{sum}}(\mathbf{h}_{i}, \mathbf{e}_{i}^{\mathrm{sum}}).
\end{equation}
The resulting representation $\mathbf{z}_{i}$ is shared by the ordinal item prediction head and the risk-level prediction head, allowing both tasks to use multimodal behavioral cues and summary-derived semantic evidence.

\subsection{Ordinal Item Prediction}
\label{subsec:item_prediction}

For each DASS item $j$, DynaBridge predicts a distribution over four ordered response levels:
\begin{equation}
\mathbf{p}_{ij}^{\mathrm{item}}
=
\operatorname{softmax}(f_j^{\mathrm{item}}(\mathbf{z}_i)),
\end{equation}
where $p_{ij}^{\mathrm{item}}(k)$ is the probability of level $k\in\{0,1,2,3\}$. The expected item score is:
\begin{equation}
\bar{d}_{ij}
=
\sum_{k=0}^{3} k \cdot p_{ij}^{\mathrm{item}}(k).
\end{equation}
Soft item predictions preserve uncertainty and provide differentiable evidence for DASS-structured reconstruction.

\subsection{DASS-Structured Item-to-Risk Reconstruction}
\label{subsec:item_to_risk}

Depression, anxiety, and stress subscale scores are reconstructed from expected item scores:
\begin{equation}
\hat{s}_{i}^{r}
=
2 \sum_{j\in\mathcal{R}_r} \bar{d}_{ij},
\quad
r\in\{D,A,S\},
\end{equation}
where $\mathcal{R}_D=\mathcal{D}$, $\mathcal{R}_A=\mathcal{A}$, and $\mathcal{R}_S=\mathcal{S}$. The factor of 2 follows the DASS-21 scoring rule. The reconstructed score is mapped to risk probability by:
\begin{equation}
p_{i,\mathrm{recon}}^{r}
=
\sigma(a_r\hat{s}_{i}^{r}+b_r),
\end{equation}
where $\sigma(\cdot)$ denotes the sigmoid function, and $a_r$ and $b_r$ are calibration parameters. In parallel, direct risk probabilities are predicted from $\mathbf{z}_i$:
\begin{equation}
p_{i,\mathrm{direct}}^{r}
=
\sigma(f_r^{\mathrm{risk}}(\mathbf{z}_i)).
\end{equation}
The final risk probability combines both branches:
\begin{equation}
p_{i,\mathrm{final}}^{r}
=
\lambda_r p_{i,\mathrm{direct}}^{r}
+
(1-\lambda_r)p_{i,\mathrm{recon}}^{r}.
\end{equation}
The fused probability $p_{i,\mathrm{final}}^{r}$ is used as the final A1 inference output, while the direct and reconstructed branches are supervised separately during training and aligned by a consistency loss. This encourages risk predictions to remain consistent with fine-grained DASS item evidence while retaining direct multimodal risk cues. The interpolation is dimension-specific because depression, anxiety, and stress have different prevalence and calibration behavior. When item predictions are reliable, the reconstructed branch strengthens psychometric consistency; when item evidence is uncertain, the direct multimodal branch preserves robustness.

\subsection{Training Objective}
\label{subsec:training_objective}

DynaBridge is optimized with risk-level supervision, item-level ordinal supervision, reconstruction supervision, and cross-task consistency:
\begin{equation}
\mathcal{L}
=
\lambda_{\mathrm{risk}}\mathcal{L}_{\mathrm{risk}}
+
\lambda_{\mathrm{item}}\mathcal{L}_{\mathrm{item}}
+
\lambda_{\mathrm{recon}}\mathcal{L}_{\mathrm{recon}}
+
\lambda_{\mathrm{cons}}\mathcal{L}_{\mathrm{cons}}.
\end{equation}
Here, $\lambda_{\mathrm{risk}}$, $\lambda_{\mathrm{item}}$, $\lambda_{\mathrm{recon}}$, and $\lambda_{\mathrm{cons}}$ are loss weights and are distinct from the risk-fusion weight $\lambda_r$. The risk and reconstruction losses are:
\begin{equation}
\mathcal{L}_{\mathrm{risk}}
=
\sum_{i,r}
\operatorname{BCE}(p_{i,\mathrm{direct}}^{r}, y_i^r),
\quad
\mathcal{L}_{\mathrm{recon}}
=
\sum_{i,r}
\operatorname{BCE}(p_{i,\mathrm{recon}}^{r}, y_i^r).
\end{equation}
For item prediction, we combine classification with an ordinal distance penalty:
\begin{equation}
\mathcal{L}_{\mathrm{item}}
=
\sum_{i,j}
\operatorname{CE}(\mathbf{p}_{ij}^{\mathrm{item}}, d_{ij})
+
\gamma
\sum_{i,j,k}
|k-d_{ij}|p_{ij}^{\mathrm{item}}(k).
\end{equation}
Here, $\operatorname{BCE}$ and $\operatorname{CE}$ denote binary and categorical cross-entropy losses, respectively, and $\gamma$ controls the strength of the ordinal-distance penalty. Direct and reconstructed risks are aligned by:
\begin{equation}
\mathcal{L}_{\mathrm{cons}}
=
\sum_{i,r}
\left\|
p_{i,\mathrm{direct}}^{r}
-
p_{i,\mathrm{recon}}^{r}
\right\|_2^2.
\end{equation}

\subsection{Inference-time Confidence-aware Item Refinement}
\label{subsec:item_refinement}

This step is applied only at inference and only to A2 item distributions. The item-level summary provides, for each DASS item, an evidence state, a confidence score, and an ordinal cue when explicit transcript-supported evidence is available. We refine an item only when the summary contains explicit evidence; items with weak, ambiguous, or unavailable evidence are not refined. For item $j$, the semantic cue is converted into a sparse distribution $\mathbf{q}_{ij}^{\mathrm{sum}}$. When the summary supports an ordinal level $\hat{d}_{ij}^{\mathrm{sum}}\in\{0,1,2,3\}$, $\mathbf{q}_{ij}^{\mathrm{sum}}$ is initialized as a one-hot distribution centered at $\hat{d}_{ij}^{\mathrm{sum}}$; otherwise, refinement is disabled.

Let $c_{ij}^{\mathrm{sum}}$ denote the confidence of the semantic evidence and $c_{ij}^{\mathrm{model}}=\max_k p_{ij}^{\mathrm{item}}(k)$ denote the neural model confidence. The summary confidence $c_{ij}^{\mathrm{sum}}$ is converted to a numerical score before thresholding. We use a conservative refinement gate:
\begin{equation}
m_{ij}
=
\mathbf{1}\!\left[c_{ij}^{\mathrm{sum}}\ge \tau_s\right]
\cdot
\mathbf{1}\!\left[
c_{ij}^{\mathrm{model}}\le \tau_m
\ \mathrm{or}\
|\hat{d}_{ij}^{\mathrm{model}}-\hat{d}_{ij}^{\mathrm{sum}}|\le 1
\right],
\end{equation}
where $\mathbf{1}[\cdot]$ denotes the indicator function, $\hat{d}_{ij}^{\mathrm{model}}=\arg\max_k p_{ij}^{\mathrm{item}}(k)$, and $\tau_s$ and $\tau_m$ are confidence thresholds. The interpolation weight is defined as $\alpha_{ij}=m_{ij}\alpha_{\max}$, where $\alpha_{\max}$ is a small upper bound. The refined distribution is then computed as
\begin{equation}
\tilde{\mathbf{p}}_{ij}^{\mathrm{item}}
=
(1-\alpha_{ij})\mathbf{p}_{ij}^{\mathrm{item}}
+
\alpha_{ij}\mathbf{q}_{ij}^{\mathrm{sum}}.
\end{equation}
When $m_{ij}=0$, we keep the original prediction, i.e., $\tilde{\mathbf{p}}_{ij}^{\mathrm{item}}=\mathbf{p}_{ij}^{\mathrm{item}}$. The refined distribution is used only for the final A2 item output and is not fed back into model training. This design prevents unsupported or conflicting LLM-derived cues from overriding confident behavioral predictions, restricting the summary branch to conservative correction rather than direct label replacement.

\begin{table*}[!t]
\centering
\caption{Comparison on the official validation set. A1 uses mean F1 as the primary metric; A2 uses mean QWK.}
\label{tab:sota_comparison}
\begin{tabular}{lcccc}
\toprule
\multirow{2}{*}{Method} & \multicolumn{2}{c}{A1: D/A/S Risk Prediction} & \multicolumn{2}{c}{A2: DASS-21 Item Prediction} \\
\cmidrule(lr){2-3} \cmidrule(lr){4-5}
 & Mean F1 $\uparrow$ & Mean AUROC $\uparrow$ & Mean QWK $\uparrow$ & MAE $\downarrow$ \\
\midrule
Official Baseline & 0.4604 & 0.7169 & 0.2675 & 0.4679 \\
CubeMLP~\cite{sun2022cubemlp} & 0.4711 & 0.7394 & 0.2778 & 0.4693 \\
Summary-enhanced Fusion~\cite{teng2026dynamic} & 0.4847 & 0.7509 & 0.2894 & 0.4692 \\
\textbf{DynaBridge} & \textbf{0.5012} & \textbf{0.7585} & \textbf{0.3216} & \textbf{0.4472} \\
\bottomrule
\end{tabular}
\end{table*}

\begin{table*}[!t]
\centering
\caption{Ablation study on the official validation set. The metadata oracle uses fields unavailable at test time and is reported only as an analytical upper bound.}
\label{tab:ablation}
\begin{tabular}{lccccc}
\toprule
\multirow{2}{*}{Method} & \multirow{2}{*}{Target} 
& \multicolumn{2}{c}{A1: D/A/S Risk Prediction} 
& \multicolumn{2}{c}{A2: DASS-21 Item Prediction} \\
\cmidrule(lr){3-4} \cmidrule(lr){5-6}
 & & Mean F1 $\uparrow$ & Mean AUROC $\uparrow$ & Mean QWK $\uparrow$ & MAE $\downarrow$ \\
\midrule
Acoustic-visual baseline & A1/A2 & 0.4604 & 0.7169 & 0.2675 & 0.4679 \\
+ School/class context & A1/A2 & 0.4747 & 0.7234 & 0.2714 & 0.4596 \\
+ Textual representation & A1/A2 & 0.4774 & 0.7374 & 0.2740 & 0.4516 \\
+ Dynamic DASS summaries & A1/A2 & 0.4847 & 0.7509 & 0.2894 & 0.4692 \\
\midrule
+ DASS-structured ordinal modeling & A2 & -- & -- & 0.3059 & 0.4657 \\
+ Item-to-risk reconstruction & A1 & 0.5012 & 0.7585 & -- & -- \\
+ Confidence-aware item refinement & A2 & -- & -- & 0.3216 & 0.4472 \\
\midrule
\textbf{DynaBridge} & A1/A2 & \textbf{0.5012} & \textbf{0.7585} & \textbf{0.3216} & \textbf{0.4472} \\
\midrule
Auxiliary metadata oracle$^\dagger$ & Oracle & 0.6468 & 0.8429 & 0.5037 & 0.3890 \\
\bottomrule
\end{tabular}

\vspace{2mm}
\footnotesize{$^\dagger$The auxiliary metadata oracle uses training/validation-only fields that are unavailable at test time. It is not used in the final submitted model.}
\end{table*}

\section{Experiments}
\label{sec:experiments}

\subsection{Dataset and Evaluation Metrics}
\label{subsec:dataset_metrics}

We evaluate DynaBridge on AdoDAS, a privacy-preserving multimodal benchmark for adolescent depression, anxiety, and stress assessment. It contains 6,000 participants and 24,000 audio-video segments. Each participant has four sessions: one fixed-text reading session and three free-response sessions about daily state, positive memory, and negative memory. Following the official subject-disjoint split, the data are divided into 4,200 training, 600 validation, and 1,200 testing participants. To protect privacy, the benchmark releases anonymized acoustic and visual representations instead of raw recordings.

Annotations are derived from DASS-21. Each of the 21 items is rated on an ordinal scale from 0 to 3 and belongs to the depression, anxiety, or stress subscale described in Section~\ref{subsec:problem}. A1 predicts binary D/A/S risks, while A2 predicts the 21 ordinal item responses. For A1, the primary metric is mean F1 over depression, anxiety, and stress; AUROC is also reported. For A2, the primary metric is mean Quadratic Weighted Kappa (QWK)~\cite{cohen1968weighted}; mean absolute error (MAE) is reported as an auxiliary metric.

\subsection{Implementation Details}
\label{subsec:implementation}

We implement DynaBridge in PyTorch~\cite{paszke2019pytorch}. Each modality is projected to a 64-dimensional adapter space and fused into a 256-dimensional hidden representation. The temporal encoder contains six residual dilated convolution layers~\cite{bai2018empirical}, and gated multiple-instance aggregation~\cite{ilse2018attention} produces participant-level representations from the four sessions.

Acoustic inputs include Mel/MFCC features and SSL speech embeddings extracted from Chinese-HuBERT-large~\cite{hsu2021hubert}. Visual inputs use SSL visual embeddings from ViT-MAE-base~\cite{he2022masked}. Transcripts and DASS summaries are encoded with \texttt{text-embedding-3-large}~\cite{openai2024embedding}. Although the fixed reading session is excluded from summary generation, its acoustic and visual features remain in the multimodal stream because they still contain paralinguistic and behavioral cues. Summary generation uses identical evidence-oriented prompts for all participants; the frozen LLM does not access DASS labels, and all summaries are cached before neural model training. For item prediction, we use four-class item heads with ordinal-distance regularization. The calibration parameters $a_r,b_r$, fusion weights $\lambda_r$, and refinement hyperparameters, including $\alpha_{\max}$, are selected on the validation split and fixed during inference.

Models are trained with AdamW~\cite{loshchilov2019decoupled}, batch size 64, initial learning rate $1\times10^{-3}$, weight decay 0.01, at most 40 epochs, early stopping, automatic mixed precision, and gradient clipping at 1.0. For ablation variants, the backbone, optimizer, and training schedule are kept unchanged; only the indicated evidence source or structural module is added or removed.

\paragraph{Reproducibility and leakage control.}
To keep the summary branch reproducible, all DASS-aware summaries are generated once before neural model training using the same evidence-oriented prompt template and fixed decoding configuration for all participants. The LLM receives only the transcripts of the three free-response sessions and never receives DASS-21 item scores, D/A/S risk labels, model predictions, validation errors, or metadata fields unavailable at test time. The fixed reading session is excluded from both text encoding and summary generation because its textual content is identical across participants; only its acoustic and visual cues are used in the multimodal stream. The generated summaries are cached and reused unchanged in all training, validation, and ablation runs.

For the final submitted system, we use only fields that are available at inference time. The auxiliary metadata oracle in Table~\ref{tab:ablation} is included only as an analytical upper bound and is not used for model selection or final prediction. Since official test labels are hidden, validation results are reported as development-set evidence rather than as a claim of hidden-test generalization. To reduce validation-specific tuning, the backbone architecture, prompt template, feature encoders, training schedule, and refinement rule are kept fixed across ablation variants; only the explicitly stated modules are added or removed.

\paragraph{Inference and output construction.}
For A1, the final risk score of each D/A/S dimension is given by the fused probability $p_{i,\mathrm{final}}^{r}$. Binary risk labels are obtained by applying fixed dimension-specific decision thresholds to these probabilities. For A2, the predicted response of each DASS item is obtained by taking the ordinal level with the maximum probability after confidence-aware refinement, i.e., $\hat{d}_{ij}=\arg\max_k \tilde{p}_{ij}^{\mathrm{item}}(k)$. When refinement is not triggered, $\tilde{\mathbf{p}}_{ij}^{\mathrm{item}}$ is identical to the original model distribution. All decision thresholds and refinement hyperparameters are fixed before final inference and are shared across all participants.

\subsection{Comparison with Representative Methods}
\label{subsec:sota_comparison}

Since test labels are not public, all results are reported on the official subject-disjoint validation split using the same evaluation protocol. We compare with the official acoustic-visual baseline, CubeMLP~\cite{sun2022cubemlp}, and a summary-enhanced fusion variant~\cite{teng2026dynamic} that uses LLM summaries but does not model the full DASS-21 cross-task structure.

Table~\ref{tab:sota_comparison} shows that DynaBridge achieves the best performance on both tracks. For A1, it improves mean F1 from 0.4604 to 0.5012 and mean AUROC from 0.7169 to 0.7585 over the official baseline. For A2, it increases mean QWK from 0.2675 to 0.3216 and obtains the lowest MAE. These gains indicate that DASS-aware summaries, ordinal item modeling, and item-to-risk reconstruction provide complementary benefits. The relative improvement is larger on A2, suggesting that the proposed DASS structure is particularly helpful for distinguishing adjacent ordinal item levels, while the A1 gain mainly comes from using item-level predictions as structured auxiliary evidence.

\subsection{Ablation Study}
\label{subsec:ablation}

Table~\ref{tab:ablation} analyzes the contribution of each module. Adding school/class context improves both A1 and A2, suggesting that participant context provides useful soft priors. Textual representations further improve A1 AUROC and A2 MAE, showing that response content complements acoustic-visual behavior. Dynamic DASS summaries bring additional gains, especially for A1 and A2 QWK, because they aggregate cross-session psychological evidence from free-response sessions.

Task-specific modules provide further improvements. DASS-structured ordinal modeling benefits A2 by preserving the ordered nature of symptom severity. Item-to-risk reconstruction improves A1 by converting item soft probabilities into D/A/S evidence through the DASS-21 scoring rule, strengthening consistency between fine-grained item responses and coarse risk labels. Confidence-aware item refinement further improves A2 by using semantic evidence only when summary confidence is high.

The auxiliary metadata oracle uses fields unavailable at test time, such as family structure, only-child status, academic performance change, and emotional state change. It is not used in the final system. Its large validation gain suggests that latent contextual factors are strongly associated with DASS outcomes, motivating our use of test-time available approximations such as context priors, dynamic summaries, and cross-task reconstruction.

\subsection{Analysis}
\label{subsec:analysis}

DynaBridge improves structured assessment for three main reasons. First, dynamic DASS summaries provide participant-level semantic evidence beyond raw transcript embeddings. By aggregating information from the daily-state, positive-memory, and negative-memory sessions, the summaries capture cross-session behavioral patterns that are difficult to represent using isolated responses alone. This provides complementary evidence to multimodal behavioral features, especially when symptom-related cues are scattered across multiple sessions.

Second, item-to-risk reconstruction explicitly aligns the two official tasks. Since A1 risk labels are derived from DASS-21 subscale scores, risk prediction should be supported by coherent item-level symptom evidence. The reconstructed branch converts A2 soft item probabilities into D/A/S risk evidence using the predefined DASS structure, while the direct branch preserves multimodal cues that may not be fully captured by item predictions. Their combination improves both psychometric consistency and robustness.

Third, confidence-aware item refinement improves A2 by incorporating transcript-supported semantic evidence conservatively. Refinement is applied only when item-level evidence is explicit and sufficiently confident, preventing unsupported LLM cues from overriding behavioral predictions. This is particularly useful for items describing specific symptoms that may be only weakly expressed in multimodal responses.

Beyond predictive performance, DynaBridge also improves inspectability. The final risk prediction can be decomposed into a direct multimodal branch and an item-derived branch, allowing high-level D/A/S risks to be traced back to fine-grained DASS item evidence. While this does not make the system clinically diagnostic, it provides a more transparent assessment process than a single end-to-end classifier.

Despite these advantages, DynaBridge remains a screening-oriented computational system rather than a clinical diagnostic tool. Its summary-guided branch depends on transcript quality, and the reported results are based on the official validation split because hidden test labels are unavailable. Future work will study cross-cohort robustness, calibration under distribution shift, and clinically grounded interpretation of fine-grained symptom predictions.

\section{Conclusion}
\label{sec:conclusion}

In this work, we presented DynaBridge, a dynamic summary-guided cross-task multimodal framework for DASS-structured mental health assessment. Instead of treating D/A/S risk prediction and DASS-21 item prediction as independent tasks, DynaBridge explicitly bridges them through ordinal item modeling and DASS-structured item-to-risk reconstruction. It further incorporates acoustic, visual, textual, and frozen-LLM-generated semantic evidence across multiple response sessions, while using confidence-aware refinement to conservatively exploit transcript-supported item-level cues. Experiments on the official validation split show consistent improvements over the official baseline and representative multimodal methods on both risk-level and item-level prediction. These results suggest that exploiting the psychometric structure of questionnaire-derived labels is beneficial for multimodal mental health assessment. DynaBridge remains a screening-oriented computational system rather than a clinical diagnostic tool. Future work will study stronger calibration under distribution shift, cross-cohort robustness, and clinically grounded interpretation of fine-grained symptom predictions.

\section{ACKNOWLEDGMENT}

S. Teng would like to thank the Program for Forming Japan's Peak Research Universities (J-PEAKS) (Grant No. R6-20) for supporting his postdoctoral position. This work was supported in part by the Grant-in-Aid for Scientific Research from the Japanese Ministry of Education, Culture, Sports, Science and Technology (MEXT) under Grant No. 20KK0234; by JSPS KAKENHI Grant No. JP23K16909; by JST CREST (JPMJCR25T4); and by the Natural Science Foundation of Zhejiang Province (Grant No. LZ22F020012).

\bibliographystyle{ACM-Reference-Format}
\bibliography{references}

\end{document}